\documentclass[11pt, a4paper, twocolumn, copyright, gr]{google}

\usepackage[authoryear, sort&compress, round]{natbib}
\uselogo{} 

\usepackage{microtype}
\usepackage{graphicx}
\usepackage{subfigure}
\usepackage{booktabs} 

\usepackage[utf8]{inputenc} 
\usepackage[T1]{fontenc}    
\usepackage{hyperref}       
\usepackage{url}            
\usepackage{booktabs}       
\usepackage{amsmath}
\usepackage{amsfonts}       
\usepackage{nicefrac}       
\usepackage{microtype}      
\usepackage[dvipsnames]{xcolor}         
\usepackage{algorithm}
\usepackage[noend]{algorithmic}
\usepackage{graphicx}
\usepackage{subfigure}
\usepackage{bbm}
\usepackage{wrapfig}
\usepackage{lipsum}

\usepackage{hyperref}

\newcommand{\setX}{\mathcal{X}}
\newcommand{\setY}{\mathcal{Y}}

\newcommand{\regret}{\mathop{\mathtt{regret}}}

\newcommand{\fcpr}{\mathop{\mathtt{PER}}}

\newcommand{\auc}{\mathtt{AUC}}
\newcommand{\tend}{t_{\text{stop}}}
\newcommand{\tmax}{{T}}

\renewcommand{\algorithmicrequire}{\textbf{Input:}}
\renewcommand{\algorithmicensure}{\textbf{Output:}}
\title{Efficient Hyperparameter Search for Non-Stationary Model Training}

\correspondingauthor{berivan@google.com}

\reportnumber{} 

\renewcommand{\today}{}

\author[1]{Berivan Isik}
\author[2]{Matthew Fahrbach}
\author[2]{Dima Kuzmin}
\author[2]{Nicolas Mayoraz}
\author[2]{Emil Praun}
\author[2]{Steffen Rendle}
\author[2]{Raghavendra Vasudeva}

\affil[1]{Google DeepMind}
\affil[2]{\thepa{}{}}

\begin{abstract}
Online learning is the cornerstone of applications like recommendation and advertising systems, where models continuously adapt to shifting data distributions. Model training for such systems is remarkably expensive, a cost that multiplies during hyperparameter search. We introduce a two-stage paradigm to reduce this cost: \textbf{(1)} efficiently identifying the most promising configurations, and then \textbf{(2)} training only these selected candidates to their full potential. Our core insight is that focusing on accurate identification in the first stage, rather than achieving peak performance, allows for aggressive cost-saving measures. We develop novel data reduction and prediction strategies that specifically \emph{overcome the challenges of sequential, non-stationary data} not addressed by conventional hyperparameter optimization. We validate our framework's effectiveness through a dual evaluation: first on the Criteo 1TB dataset, the largest suitable public benchmark, and second on an industrial advertising system operating at a scale two orders of magnitude larger. Our methods reduce the total hyperparameter search cost by up to 10$\times$ on the public benchmark and deliver significant, validated efficiency gains in the industrial setting.
\end{abstract}

\begin{document}

\maketitle

\section{Introduction} \label{sec:introduction}
Online learning~\citep{hoi2021online} is a common training strategy, crucial for applications like recommendation~\citep{naumov2019deep} and advertising~\citep{mcmahan2013ad, richardson2007predicting} systems, where models continuously train on new, sequentially available data to adapt to distribution shifts over time, e.g., time-series logs of user interaction with new items, new ads, or new feedback. These systems, which form the backbone of platforms like YouTube, Facebook, and Google Ads, must adapt to evolving user preferences and item popularities. While models are continuously updated during deployment, they must periodically be replaced by stronger versions with architectural changes or improved hyperparameters. This replacement process is guided by an extensive search over candidate configurations, trained on fixed historical data that preserves the sequential ordering of live traffic—a process known as backtesting.

 The hyperparameter search process is prohibitively costly, as numerous candidate configurations must be trained on massive datasets. For recommendation and ads models, the cost is particularly high due to large volume of data needed where training datasets can reach hundreds of billions of examples~\citep{anil2022factory, jain2024data, kurian2025scalable}. The core scientific challenge compounding this cost is the non-stationary nature of the data. Unlike traditional offline learning, which often assumes independently and identically distributed (i.i.d.) data, online learning systems contend with constant distribution shift, where the underlying data-generating process changes over time. This temporal shift renders many conventional hyperparameter optimization (HPO) techniques, such as those relying on smooth, monotonic learning curves, fundamentally unsuitable. This elevates the problem from a practical engineering challenge to a fundamental research question in machine learning under distribution shift.   

This work introduces a two-stage paradigm to reduce the cost of hyperparameter search in non-stationary online learning environments. The first stage efficiently identifies the most promising configurations, while the second stage trains only these selected candidates to their full potential. The key insight is that the first stage does not require candidates to achieve their best possible performance; its sole objective is to accurately identify the top configurations. This allows for aggressive cost-saving strategies. 

A suitable benchmark for this study must satisfy two stringent criteria: \textbf{(1) massive scale}, with billions of examples, to accurately reflect the computational costs and data complexities of industrial systems, \textbf{(2) fine-grained, sequential, time-ordered data} spanning a significant duration to capture the non-stationary dynamics and temporal distribution shifts inherent to production environments. As will be detailed in Section 4.1.1, the Criteo 1TB  dataset~\cite{criteo_data} is, to the best of our knowledge, the only publicly available resource that satisfies both of these critical requirements.
 
The contributions of this work are as follows:

\begin{itemize}
    \item Novel data reduction and prediction strategies are developed for efficient hyperparameter search. These include a performance-based stopping mechanism that dynamically terminates unpromising training runs and advanced prediction techniques (trajectory prediction and stratified prediction) designed to forecast final performance from partial training data under distribution shift.

    \item A comprehensive evaluation of these methods is conducted on the Criteo 1TB benchmark, demonstrating up to a 10x reduction in the computational cost required to identify top-performing configurations.

    \item The practical impact and scalability of the proposed framework are validated on a \textbf{real-world, industrial advertising system operating at a scale two orders of magnitude larger than the Criteo benchmark}, where it achieves significant efficiency gains. This dual validation underscores both the scientific novelty and the industrial relevance of the contributions.
\end{itemize}
\section{Related Work} \label{sec:related_work}

\paragraph{Recommendation and Advertising Systems. }  Recommendation and advertising systems form the backbone of many online platforms, such as YouTube~\citep{covington2016deep}, Facebook ads~\citep{he2014practical}, Bing search~\citep{ling2017model}, Google ads~\citep{anil2022factory,coleman2024unified,kurian2025scalable,mcmahan2013ad}, TikTok~\citep{liu2022monolith}, and Amazon recommendations~\citep{linden2003amazon}. Online learning is particularly relevant in these applications due to the constant influx of new data and the need to adapt to evolving user preferences and item popularity \cite{anil2022factory, mcmahan2013ad}. This inherent distribution shift in online data introduces unique challenges, necessitating distinct solutions compared to standard ``offline'' learning approaches that assume distributionally stationary data~\citep{bedi2018tracking,fahrbach2023learning, hazan2007adaptive, yang2016tracking, zinkevich2003online}. Our work addresses one such challenge: efficient hyperparameter search. While this problem has been extensively studied for offline learning, the complexities introduced by the distribution shift demand new analyses and tools. 

\paragraph{Data-Efficient Training. } The increasing scale of modern datasets has driven significant interest in data-efficient learning techniques. These methods aim to achieve high performance with limited, selectively chosen subset of the full training data. One line of work focuses on selecting weighted subsets of data that best approximate the full gradient~\citep{killamsetty2021grad,killamsetty2021glister,  mirzasoleiman2020coresets, pooladzandi2022adaptive, yang2023towards}. Another approach leverages ``importance signals'', such as (expected) gradient norm~\citep{alain2015variance, katharopoulos2018not, paul2021deep}, per-example loss~\citep{jiang2019accelerating, loshchilov2015online}, and forgettability~\citep{toneva2018an},  to guide data reduction by retaining the most informative examples. This selection process can occur after some steps of training or after a full training run on some cheaper proxy model~\citep{coleman2019selection}. While data-efficient training is a crucial component of our proposal, existing methods do not address the distribution shift inherent in online learning. Consequently, data-efficient online learning has relied on simpler strategies like uniform or label-dependent sub-sampling of examples at more conservative rates~\citep{jain2024data}. Our work overcomes this limitation by enabling  a successful hyperparameter search even if the candidate configurations do not maintain their best performance on aggressively reduced data. This allows for the exploration of more aggressive data reduction strategies, which, when supported by our prediction strategies, facilitates significantly more efficient hyperparameter search. 

\paragraph{Hyperparameter and Architecture Search. } Hyperparameter and neural architecture search  are critical but computationally expensive components of the machine learning lifecycle. Traditional hyperparameter search methods include grid search, random search~\citep{bergstra2012random}, and Bayesian optimization~\citep{snoek2012practical}. More recent approaches leverage techniques like evolutionary algorithms~\citep{real2017large} and gradient-based optimization \cite{liudarts}. Neural architecture search aims to automate the design of neural network architectures, often employing search strategies similar to those used in hyperparameter search~\citep{elsken2019neural, zoph2022neural}. While these methods have shown considerable success, they often require substantial computational resources. This has motivated research in efficient hyperparameter search, aiming to identify the top configurations with reduced cost~\citep{feurer2019hyperparameter, gao2022efficient, shen2022efficient}. A common technique adopted by practitioners is to stop the training runs of non-promising configurations early (aka \emph{early-stopping}) based on their performance during the initial stages. This strategy can also rely on predicting the configurations' full learning trajectory using Bayesian optimization~\citep{chandrashekaran2017speeding, klein2017learning, lin2025successive}, autoregressive
models~\citep{wistuba2020learning}, or scaling laws~\citep{domhan2015speeding, kadra2023scaling}. However, these prediction techniques are most effective when loss curves are smooth and monotonic. In online learning, distribution shifts in the data cause performance to fluctuate, making such predictions highly challenging. Our work directly addresses this challenge by mitigating the effect of distribution shift with a reference model and by aggregating ``sliced'' performance predictions over more stable data clusters.  

\paragraph{Early Stopping and Successive Halving. } A critical component of efficient hyperparameter search is the early termination of unpromising configurations. A dominant paradigm in this area is bandit-based adaptive resource allocation, with Successive Halving (SHA) serving as a foundational algorithm~\citep{jamieson2016non, kumar2018parallel, soper2022hyperparameter, triepels2023sasha, li2020system, salinas2021multi, schmucker2021multi}. SHA operates by allocating an initial budget to a large set of randomly sampled configurations. After this initial evaluation, it discards half of the worst-performing configurations and reallocates the budget to the survivors for further training. This process of evaluation, promotion, and pruning is repeated in successive ``rungs'' until only one configuration remains. SHA is a theoretically principled method that effectively allocates more resources to more promising configurations.  However, SHA's performance is sensitive to the trade-off between the number of configurations evaluated and the minimum budget allocated to each~\citep{egele2024unreasonable}. This is known as the ``n vs. r'' trade-off: allocating a small initial budget allows for exploring many configurations but risks prematurely eliminating ``late bloomers'', while a large initial budget is more robust but limits the number of configurations that can be explored. Hyperband~\citep{li2018hyperband} was introduced as a meta-algorithm to address this dilemma. It executes multiple brackets of SHA, each with a different initial resource allocation. By hedging across various exploration-exploitation trade-offs, Hyperband robustly and automatically adapts to the problem at hand without requiring prior knowledge of convergence rates, often achieving significant speedups over other hyperparameter search methods like Bayesian optimization.

\paragraph{Positioning Our Work. } Some of our strategies build upon and generalize this established line of work on SHA. The proposed performance-based stopping mechanism (Algorithm~\ref{algo:adaptive_stopping} in Section~\ref{sec:method}) is a generalized SHA framework. Specifically, the variant of this framework that uses ``constant prediction'' (see Section~\ref{sec:method} for details),  where the performance at an early step is used as a proxy for final performance, is a direct generalization of the SHA algorithm. Whereas SHA typically employs a fixed reduction factor of $
\eta=2$, our framework allows for a flexible stopping ratio $\rho$, where $\rho = 1 - 1/{\eta}$, making it more adaptable to different search problems and resource constraints.   

While this generalized SHA serves as a strong baseline, the core novelty of our work lies in moving beyond simple performance extrapolation to \textbf{address the unique challenges of non-stationary data}. Standard SHA and Hyperband are most effective when learning curves are relatively well-behaved. However, in online learning, distribution shifts cause performance metrics to fluctuate, producing non-monotonic trajectories that violate the underlying assumptions of many early-stopping and SHA methods. The proposed trajectory prediction and stratified prediction strategies are designed specifically to handle these complex learning dynamics, a challenge not explicitly addressed by standard SHA or Hyperband. These advanced methods mitigate the effects of distribution shift by modeling performance trajectories and disaggregating predictions across more stable data slices, thereby enabling more accurate and robust early-stopping decisions in non-stationary environments.

\section{Problem Setup and Preliminaries} \label{sec:problem}

\subsection{Problem Setup}
Let $\setX$ be the space of input features and $\setY$ the space of labels.
A tuple $(x,y) \in \setX \times \setY$ is a training example consisting of input features $x$ and the associated label $y$.
For example, $x$ could contain information about a query and a product, and the label $y$ could indicate whether or not the product was clicked for the query. Let $f(x, \theta, \omega)$ be a machine learning model that predicts the label (or a statistic of the label, e.g., its probability) for a given example $x$.
Here $\theta \in \Theta$ are the \emph{parameters} of the model that are trained and $\omega \in \Omega$ is a \emph{configuration}.
We use a very broad notion of configuration, which can include model structure such as Cross Networks~\citep{wang2017deep, wang2021dcn} and  Factorization Machines~\citep{rendle2010factorization}, architectural parameters such as embedding dimension, optimization parameters such as learning rate schedule and batch size, and data processing parameters, such as the vocabulary size.

\paragraph{Performance Metrics. } We use the term \emph{performance metric} to refer to metrics that measure the performance of a \emph{single} model configuration for predicting labels $\setY$, such as log loss. Later, we will define \emph{ranking metrics} that measure how well a \emph{set} of model configurations is ordered based on the configurations' \emph{performance} metrics. 
Let $m$ be a performance metric that compares one or several predictions of a model with the labels, such as log loss or area under the curve ($\auc$).
Without loss of generality we assume all performance metrics are loss metrics, i.e., smaller the better.
We can translate any quality metric to a loss metric by taking the negative value of the quality. 

\paragraph{Online Training and Evaluation. } Let $(x_1,y_1), (x_2, y_2), \ldots, (x_\tmax, y_\tmax)$ be a sequence of training examples.
Online training is an iterative training process where at any time step $t$, the model parameters $\theta_t$ of a configuration are only influenced by data $(x_1,y_1),\ldots,(x_t,y_t)$.
Likewise in online evaluation, we obtain a series of metrics $m_1, m_2, \ldots$, where $m_t$ is computed based on the model parameters $\theta_{t-1}$ from a time before $t$.
We define the average metric over a set $\mathcal{W}$ of points in time of a configuration $\omega$ as $\overline{m}_{\mathcal{W}} := \frac{1}{|\mathcal{W}|} \sum_{t \in \mathcal{W}} m_t$.
Ultimately, we are interested in the average metric over an evaluation time window over the most recent $\Delta+1$ time steps: $\mathcal{W}_{\text{eval}}=[\tmax - \Delta,\tmax]$. (We use $[a,b]:=\{i \in \mathbb{N} : a\leq i \leq b\}$ for closed intervals over natural numbers.)
For convenience, we define $\overline{m} :=\overline{m}_{[\tmax-\Delta,\tmax]}$. 

\paragraph{Our Goal: Ranking Configurations. } Our goal is to rank configurations. 
Let $r : [|\Omega|] \rightarrow \Omega$ be an ordering of $\Omega$.
Let the ranking with respect to metric $m$ be $r$ such that $\overline{m}(r(i)) < \overline{m}(r(j)) \implies i < j$. We denote by $r^*$ the ground truth ranking associated with $\overline{m}$ that are computed on configurations trained on the full data. We aim to efficiently compute a ``good'' ranking $r$ that is close to $r^*$.
Specifically, we want to avoid expensive training of all candidate configurations on the full data for computing the metrics $\overline{m}$. We let $C$ denote the ratio of the computational cost needed to obtain a ranking $r$ to the cost of obtaining $r^*$ by training all configurations on the full data ($C = \frac{\text{cost of obtaining } r}{\text{cost of obtaining } r^*}$).
Next, we discuss how to measure the ``goodness'' of a ranking $r$ with respect to the ground truth ranking $r^*$. 

\subsection{Ranking Metrics: Comparing the Predicted Ranking $r$ and Ground Truth Ranking $r^*$ }
 
\paragraph{Pairwise Error Rate ($\fcpr$).}\footnote{This is also the negative of ROC AUC -- used for evaluating binary classification models.} Given a ranking $r$, we can find the percentage of incorrectly ranked configuration pairs among all possible pairwise configuration comparisons based on their ground truth comparisons in $r^*$, using the following formula: $\fcpr(r) = \frac{1}{\frac{1}{2}|\Omega| (|\Omega|-1)}
    \sum_{i=1}^{|\Omega|} \sum_{j=i+1}^{|\Omega|}
    \mathbbm{1}\{\overline{m}(r(i)) > \overline{m}(r(j))\}$, 
where $\mathbbm{1}\{P\} = 1$ when $P$ is true and $0$ otherwise.

\paragraph{Regret.} $\fcpr$ measures how accurately we rank the configurations and penalize misrankings in a discrete sense. For instance, $\fcpr$ penalizes a configuration pair where one is only slightly better than the other the exact same way as a pair where one is much better than other, when ranked inaccurately. However, in reality, misranking is a bigger problem when there is a significant difference between the configurations' performance. To capture this, we use another ranking metric, $\regret$, defined as: $\regret(r) = \frac{1}{|\Omega|} \sum_{i=1}^{|\Omega|} \max\left(0, \overline{m}(r(i))- \overline{m}(r^*(i)) \right).$ $\regret$ measures how much of a difference misrankings in $r$ make in terms of performance metrics. This also allows us to set an ``acceptable'' $\regret$ level at the variance of $\overline{m}$ due to randomness in the training (more details in Section~\ref{sec:experiments}).   

Both $\fcpr$ and $\regret$ reflect what fraction of pairs among \emph{all} pairs of configurations in a given pool are misranked. However, in practice, hyperparameter search only cares about identifying and ranking the \emph{top} configurations. For this purpose, $\fcpr$ and $\regret$ are not the most indicative ranking metrics. For instance, one method can be worse than another in $\fcpr$ or $\regret$ but be much better in ranking the top-$k$ configurations, which is what matters for a successful hyperparameter search. We now introduce our \textbf{main metric}, $\regret @k$, that specifically measure how well we predict the ranking of top-$k$ configurations, allowing for mistakes in the remaining configurations: $    \regret@k(r) = \frac{1}{k} \sum_{i=1}^{k} \max\left(0, \overline{m}(r(i))- \overline{m}(r^*(i)) \right).$ $\regret@k$ measures how much additional performance metric loss we would obtain if we used the top-$k$ configurations in $r$ instead of the top-$k$ configurations in the ground truth ranking $r^*$. 

\begin{figure}[h]
  \centering
  \includegraphics[width=\columnwidth]{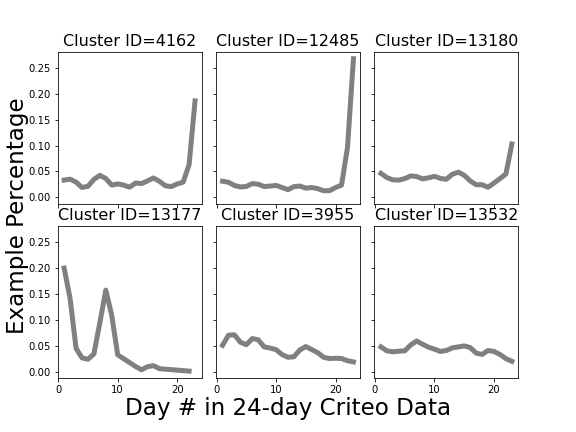}
  \caption{Cluster sizes show high variation over the 24 days of training window. }
  \label{fig:cluster_ratio}
\end{figure}

\begin{figure*}[h]
        \centering
        \includegraphics[width=0.42\linewidth]{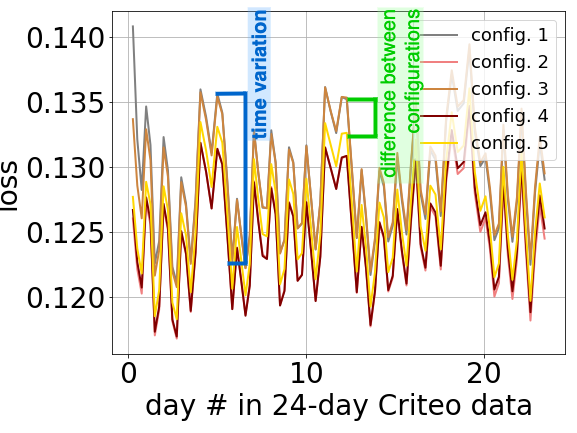}
        \includegraphics[width=0.42\linewidth]{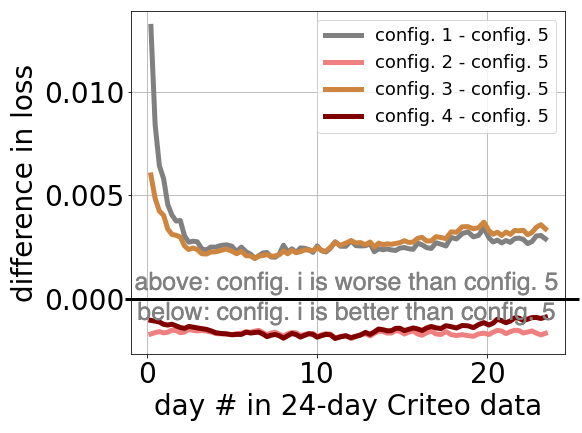}
        \caption{\textbf{(left)} The effect of time variation in sequential non-stationary data on the loss during training. Different configurations (architecture, model size, and other hyperparameters are varied) are all trained on the same 24-day Criteo data and they follow the same time variation pattern. The time variation (e.g. \textcolor{blue}{blue} line) is significantly higher than the difference between configurations (e.g. \textcolor{ForestGreen}{green} line). \textbf{(right)} Relative loss with respect to a reference configuration. We choose Configuration 5 from (left) as a reference run and plot the other configurations' loss with respect to that.}
        \label{fig:time variation}
\end{figure*}

\subsection{Time Variation in Online Training Data}
\label{sec:time_variation}

\paragraph{Feature distribution changes over time.} The time variation in feature distribution is one of the unique challenges we face in hyperparameter search for online learning~\citep{fahrbach2023learning}. To better understand this variation, we group the Criteo dataset into $15,000$ clusters with a $k$-means clustering on the embeddings of each example. These embedding are obtained from a Variational Autoencoder (VAE) model we design and train to cluster the Criteo data. In addition to demonstrating the distribution shift in the data, these clusters also play an important role in some of our data reduction strategies that we introduce in Section~\ref{sec:stratified}. As training progresses over the 24-day Criteo data, we track how the cluster distribution (fraction of data in each cluster) changes over time. Figure~\ref{fig:cluster_ratio} visualizes this change for a selected set of clusters. It is seen that some clusters do not have much data until the last few days of training but there is a significant increase in their size in the late stages of training. For some other clusters, we see the opposite trend. This demonstrates the high variation in the feature distribution, which makes predictions for later days, e.g., the evaluation window, hard.

\paragraph{Time variation in loss is more dominant than the difference between two candidate models.} Figure~\ref{fig:time variation}(left) shows how the loss progresses over the course of online training on 24-day Criteo data. The distribution shift in training data introduces a time variation in the loss trajectory. For Figure~\ref{fig:time variation}-(left), the five configurations include two Factorization Machine (FM) models, two Cross Network (CN) models, and one Mixture of Experts (MoE) model, with different model size and optimization hyperparameters. Notice the significantly larger time variation in the loss of an individual configuration, shown in \textcolor{blue}{blue}, compared to the difference between different configurations, shown in \textcolor{ForestGreen}{green}. Such small difference between configurations compared to the variation of an individual configuration makes the ranking task challenging.

\paragraph{Time variation in loss is consistent across candidate models.}  Figure~\ref{fig:time variation}(left) also shows that time variation in the loss follows a similar pattern across all the configurations even though these configurations have different architectures and hyperparameters -- suggesting that the time variation pattern is attributed to a "problem hardness" aspect inherent in the data and shared across all configurations. 
\paragraph{Dealing with time variation in ranking candidate models.} While high time variance in sequential non-stationary data makes ranking configurations challenging, the consistent pattern across different architectures provides an opportunity to mitigate the effect of time variation: We take the \emph{difference} between the configurations' performances, rather than using absolute metrics directly. As demonstrated in Figure~\ref{fig:time variation}(right), this approach significantly reduces the overall variance and mitigates the impact of distribution shift. Therefore, in next sections, we primarily rely on this relative performance.

\section{Method} \label{sec:method} 
In this section, we introduce our data reduction and prediction strategies. 

\subsection{Data Reduction Strategies}
\subsubsection{Stopping Training Runs of Candidate Configurations}

\paragraph{One-Shot Early Stopping.} Instead of training the candidate configurations on the whole sequence $(x_1,y_1), \ldots, (x_\tmax,y_\tmax)$ and measuring the quality $\overline{m}_{[\tmax-\Delta,\tmax]}$ on the evaluation period, we stop training at time $\tend < \tmax$ and use the metrics obtained up to $\tend$ to produce a ranking.
In the simplest case, we build the ranking $r$ based on $\overline{m}_{[\tend-\Delta,\tend]}$. We term this ``constant prediction'' and provide more details in Section~\ref{sec:method_prediction} -- together with  more advanced prediction methods.
The relative cost of ranking configurations with one-shot early stopping is $C(\tend) = \frac{\tend}{\tmax}$. This relative loss decreases further when it is combined with other data reduction strategies.

\begin{algorithm*}[h]
    \caption{Performance-Based Stopping}\label{algo:adaptive_stopping}
    \begin{algorithmic}[1]
    \REQUIRE configurations $\Omega$, stopping steps $\mathcal{T}_{\text{stop}} \subseteq \{1, \dots, \tmax\}$, ratio $\rho$ of stopped configurations at any stopping step
    \ENSURE ranking $r$ 
    \STATE $\Omega_{\text{remaining}} \gets \Omega$
    \STATE Initialize ranking sequence $r \gets []$. 
    \WHILE{$\Omega_{\text{remaining}} \ne \emptyset$ and $t \leq \tmax$}
        \STATE Incremental training with $(x_t,y_t)$.
        \IF{$t \in \mathcal{T}_{\text{stop}}$}
            \STATE $\hat{m}_{\text{remaining}} \gets \textsc{PredictPerformance}(\Omega_{\text{remaining}}, t, \tmax)$
            \STATE $r_{\text{remaining}} \gets \textsc{RankConfigurations}(\Omega_{\text{remaining}}, \hat{m}_{\text{remaining}} )$
            \STATE $\Omega_{\text{pruned}}, r_{\text{pruned}} \gets \text{last } \rho \cdot |\Omega_{\text{remaining}}| \text{ configurations in } r \text{ and their rankings}$
            \STATE $r \gets \text{concatenate}(r_{\text{pruned}},r)$
            \STATE $\Omega_{\text{remaining}} \leftarrow \Omega_{\text{remaining}} \setminus \Omega_{\text{pruned}}$
        \ENDIF
    \ENDWHILE
    \STATE $m_{\text{remaining}} \gets \textsc{ComputePerformance}(\Omega_{\text{remaining}})$
    \STATE \textbf{return} $\text{concatenate}(\textsc{RankConfigurations}(\Omega_{\text{remaining}}, m_{\text{remaining}}), r)$
    \end{algorithmic}
\end{algorithm*}

\paragraph{Performance-Based Stopping.} Usually, we are not interested in the exact ranking of configurations that perform poorly.
Instead, we want to identify and rank the \emph{top} configurations.
Thus, instead of early stopping \emph{all} configurations after the \emph{same} number of steps, it is beneficial to observe the trajectory of the metrics $m_1, m_2, \ldots$, stop some non-promising training runs earlier based on their (predicted) performance, and let promising runs continue longer.
This resource allocation across candidate configurations saves costs on poor configurations that we can invest in more accurate rankings of more promising candidates.
Algorithm~\ref{algo:adaptive_stopping} sketches this idea.   

Given a set of stopping steps $\mathcal{T}_{\text{stop}} \subseteq \{1, \dots, \tmax\}$ and a stopping ratio $\rho$, performance-based stopping (1) pauses training at each stopping time $\tend \in \mathcal{T}_{\text{stop}}$, (2) predicts the performance of each configuration that is not already stopped at an earlier stopping time $\tend'<\tend$, (3) ranks these $n_{\text{remaining}}$ configurations based on the predicted performances, (4) stops training of the $\rho \cdot n_{\text{remaining}}$ least promising configurations, and (5) continues training the remaining $(1-\rho) \cdot n_{\text{remaining}}$ configurations. It can utilize any of the prediction strategies described in Section~\ref{sec:method_prediction}. We note that, successive halving, developed for similar optimization problems~\citep{kumar2018parallel, soper2022hyperparameter}, is a special case of this strategy when the prediction strategy is ``constant prediction'' and $\rho=\frac{1}{2}$. The relative cost of performance-based stopping, as a function of $\mathcal{T}_{\text{stop}}$ and $\rho$, is $C(\mathcal{T}_{\text{stop}}, \rho) =  \frac{1}{T}  \sum_{t_i \in (\mathcal{T}_{\text{stop}} \cup \{T\})} (1-\rho)^{i-1}(t_i - t_{i-1}) $.

\subsubsection{Data Sub-Sampling}

So far, we have covered methods that train on all data up to a certain time step.
A further technique to reduce training time, that is orthogonal to the other data reduction strategies, is to skip some training examples.
We use uniform sampling and label-dependent sampling (where the majority class, e.g., negative labels, are sub-sampled) of training examples as a source of additional data reduction.
Let $\lambda_y$ be the fraction of sub-sampled examples from class $y$.
Then, the relative cost as a function of $\lambda = \{\lambda_y \}_{y \in \mathcal{Y}}$ would be $C(\lambda) = \frac{1}{\tmax}\sum_{t=1}^\tmax \lambda_{y_t} = \frac{1}{\tmax}\sum_{y \in \mathcal{Y}} (\text{\# of examples from class } y) \cdot \lambda_{y}$, where $\mathcal{Y}$ is the set of all classes.

\subsection{Prediction Strategies}
\label{sec:method_prediction}
In the previous section, we have introduced data-reduction strategies that produce only parts of the metrics, e.g., by stopping training early. This means they do not allow precisely computing $\overline{m}_{[\tmax-\Delta,\tmax]}$ and ranking the configurations based on that.
We now describe prediction strategies to estimate $\overline{m}_{[\tmax-\Delta,\tmax]}$ from metrics that have been measured up to $\tend$.
We denote these estimators $\hat{m}_{\tend}$.

\subsubsection{Constant Prediction}
Constant prediction assumes that the performance seen at some point $\tend$ early in training is a good indicator of the final performance of a candidate on the evaluation data, i.e., $\hat{m}^{\text{constant}}_{\tend} = \overline{m}_{[\tend-\Delta,\tend]}$. 

\subsubsection{Trajectory Prediction} \label{sec:scaling_laws}

 To predict a configuration's trajectory, we fit a parameterized law $f$ to its observed performance up to some training point $\tend$, using the data fraction $D = \frac{\tend}{\tmax}$ as a parameter. We then use the optimized parameters to extrapolate the configuration's full-training performance, i.e., $\hat{m}^{\text{trajectory}}_{\tend} = f(1)$. A typical example for $f$ is the inverse power law, $f(D) := E + \frac{A}{D^\alpha}$, where $E$, $A$, and $\alpha$ are scalars. To reduce prediction variance, as noted in Section~\ref{sec:time_variation}, we jointly optimize these laws across all configurations by minimizing the fit error on their pairwise performance differences, instead of their absolute metrics. This optimization minimizes an objective function of the form: $\sum_{\omega, \omega' \in \Omega} \sum_{\text{a few } t \in \mathcal{T}_{\text{stop}}} \big ( (f_{\omega}(\frac{t}{\tmax}) - f_{\omega'}(\frac{t}{\tmax})) - \overline{m}_{\omega-\omega',[t-\Delta,t]}  \big )^2$, where $\overline{m}_{\omega-\omega',[t-\Delta,t]}$ denotes the average performance difference between configuration $w$ and $w'$ in time window $[t-\Delta,t]$, and $f_{\omega}$ is the law for $\omega$, dependent on its parameters $E_{\omega}$, $A_{\omega}$, and $\alpha_{\omega}$. When fitting $f$'s, an aggregation window smaller than the full evaluation window size $\Delta$ can also be used.

\subsubsection{Stratified Prediction} \label{sec:stratified}
Recall from Section~\ref{sec:time_variation} that different clusters within the training data follow a different distribution shift. While constant and trajectory prediction make a single \emph{aggregate} prediction and thus overlook this varying distribution shift across clusters, our stratified prediction strategy accounts for it through ``sliced predictions'' followed by aggregation. Let $\mathcal{S}=\{S^{(l)}\}_{l=1}^{L}$ be a partitioning of the training examples from the interval $[1,\tmax]$ into $L$ slices and $\hat{m}^{(l)}_{\tend}$ be the predicted performance on data from slice $S^{(l)}$. We assume that the model's performance on a slice is roughly independent of other slices. To make a prediction for a candidate configuration's performance $\overline{m}_{[\tmax-\Delta,\tmax]}$ on the evaluation period, we break the sum over examples into sub-sums corresponding to each slice:  

\begin{equation}
\begin{aligned}
& \overline{m}_{[\tmax-\Delta,\tmax]} = \frac{1}{\Delta+1} \sum_{t=\tmax-\Delta}^{\tmax} m_t = \\ 
& \frac{1}{\Delta+1} \sum_{S^{(l)} \in \mathcal{S}} \left[ \sum_{t=\tmax-\Delta}^{\tmax}  \mathbbm{1}\{x_t \in S^{(l)}\} \cdot m_t\right].
\end{aligned}
\end{equation}
We then predict the performance on each slice as $\hat{m}^{(l)}_{\tend}$ and reweigh them by the number of evaluation examples (from the interval $[T-\Delta, T]$) in the slice to obtain the final prediction $\hat{m}_{\tend}^{\text{stratified}}$:
\begin{align}
\hat{m}_{\tend}^{\text{stratified}} &= \frac{1}{\Delta+1} \sum_{{S^{(l)}} \in \mathcal{S}} \left[ \hat{m}_{\tend}^{(l)} \cdot \sum_{t=\tmax-\Delta}^{\tmax}  \mathbbm{1}\{x_t \in S^{(l)}\}\right],
\end{align}
where $\hat{m}^{(l)}_{\tend}$ is either the constant or trajectory prediction for slice $S^{(l)}$, with the data available for the slice at time $\tend$. In our experiments, we partition the training data into slices based on the examples' cluster assignments. We provide more details on clustering in Section~\ref{sec:experiments}. 
\section{Experimental Results} \label{sec:experiments}
This section presents a rigorous evaluation of the proposed methods, first through a comprehensive study on a large-scale public benchmark and second through validation on a web-scale industrial system.

\subsection{Experiments with a Public Benchmark Dataset}

\subsubsection{Setup} 
\paragraph{Dataset. } For a valid and realistic evaluation of our work on hyperpatameter search for non-stationary online learning, we need a dataset that exhibits both massive scale, to reflect industrial computational challenges, and a continuous, time-ordered data stream, to enable the study of temporal distribution shift. A systematic review of available datasets reveals that the Criteo 1TB predicting the click-through rate (pCTR) dataset~\citep{criteo_data} is uniquely suited for this task. Classic recommender datasets like MovieLens~\citep{harper2015movielens} and Netflix~\citep{bennett2007netflix}, while historically significant, are orders of magnitude too small for web-scale research. More importantly, their coarse temporal granularity (often daily timestamps) is insufficient to model the fine-grained sequential dynamics of online learning systems where distribution shifts can occur on much shorter time scales. Other large-scale datasets, such as Amazon Reviews~\citep{amazon1, amazon2, amazon3, amazon4}, focus on explicit feedback (reviews) rather than the implicit clickstream data common in advertising. While other advertising datasets like Avazu~\citep{avazu} exist, they are substantially smaller or do not provide a continuous, multi-week data stream necessary for studying long-term non-stationarity. 

We note that we use the Criteo 1TB dataset~\citep{criteo_data} that is 100 times larger than the commonly used Criteo pCTR dataset~\cite{criteo-display-ad-challenge}. This dataset has a sub-sampled portion of Criteo's traffic over 24 days. Each data point corresponds to a display ad served by Criteo, where the clicked ads are associated with the positive label and the others are associated with the negative label. The ads in the Criteo dataset are chronologically ordered -- suitable for the continuously evolving data setup  used for recommendation and ads models. The same dataset serves as a test set as well: during the single pass over the data, the performance metric computed at iteration $t$ is recorded as the test metric and then used to update the model parameters. Our target evaluation period is the last 3 days of the Criteo dataset, i.e., $\Delta=\text{3 days}$. We choose this evaluation window as it captures the time variation across multiple days during evaluation but also allows the model to learn the time variation across 3 weeks in the first 21 days of the training data. 

\begin{figure*}[h]
        \centering        \includegraphics[width=\linewidth]{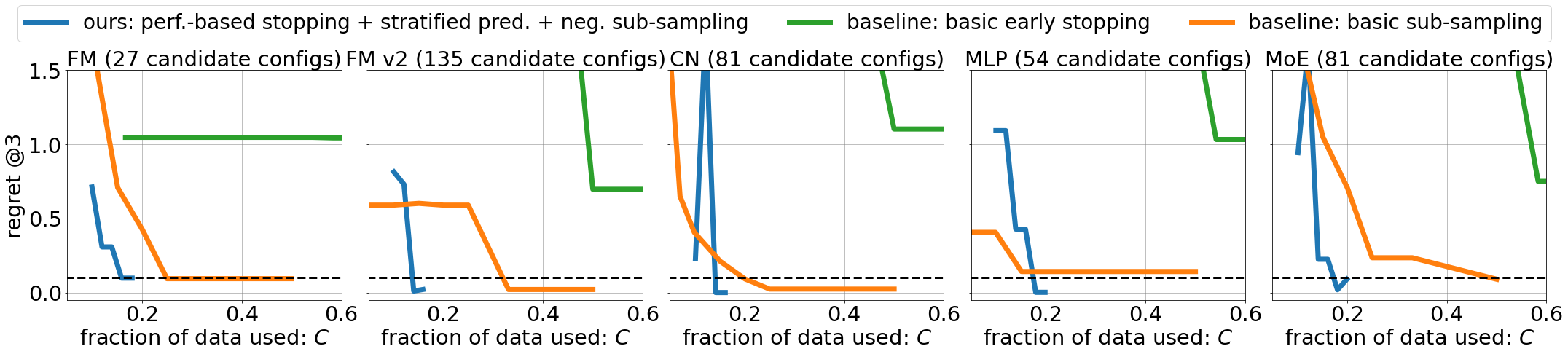}
        \caption{Our proposal (performance-based stopping with stratified prediction on sub-sampled data) in comparison with baselines, (1) basic early stopping and (2) basic sub-sampling. For our proposal, the sub-sampling is for negative-labeled examples only at a fixed rate of $0.5$. For each curve, we vary certain parameters to obtain $\regret @3$ at different $C$, e.g. $\mathcal{T}_{\text{stop}}$ for performance-based stopping (\textcolor{Blue}{blue}), $\tend$ for basic early stopping (\textcolor{ForestGreen}{green}), and $\lambda^{\text{uniform}}$ for basic sub-sampling (\textcolor{Orange}{orange}).}
        \label{fig:combined_method}
\end{figure*}

\paragraph{Candidate Configurations. } As the model architecture, we use  Factorization Machines (FMs)~\cite{rendle2010factorization}, Cross Networks (CNs)~\citep{wang2017deep, wang2021dcn}, MLP, and Mixture of Experts (MoEs)~\citep{shazeer2017outrageously}. For each architecture, we explore a variety of configurations by sweeping the optimization parameters such as learning rate, weight decay, and final learning rate. We also vary some architectural parameters, such as feature embedding dimensions and number of layers. 

\paragraph{Baselines. } While there is no prior work on predicting ranking of models in an online learning setting, we choose uniform sub-sampling and one-shot early stopping with constant prediction as baselines since these are rather straightforward approaches that could be used by practitioners, and refer to them as \textbf{basic sub-sampling} and \textbf{basic early stopping}. 

\paragraph{Clustering. }  For stratified prediction, we group the Criteo training examples into $15,000$ clusters with a $k$-means clustering on examples' embeddings extracted from a proxy model.  The proxy model is a combination of a VAE with a pCTR predictor based on High Order Factorization Machines (HOFM). We insert a bottleneck layer of dimension $32$ before the last layer of the HOFM predictor, and use the embeddings from this layer for clustering. We then group the clusters with similar distribution shift together and create the slices used in stratified prediction. This can be done in different ways -- we do this grouping at each stopping time $\tend$, based on cluster sizes. 

We only present $\regret @3$ results here due to page limit; other metrics are provided in Appendix~\ref{sec:app_exp_results}.

\subsubsection{Results}
\paragraph{Inherent model variance guides the acceptable $\regret @k$ level. } Our main metric, $\regret @k$, quantifies the ``performance loss'' incurred by misrankings among the top-$k$ confgurations. To establish an acceptable level for this performance loss, we conduct a sensitivity analysis.  First, for interpretability, we normalize $\regret @k$ by a reference model's average performance metric over the evaluation window. The reference model is trained on full data and, in practice, can be chosen as the previously deployed model. Second, we measure the sensitivity of the performance metrics to model initialization randomness across $8$ seeds. Our analysis reveals that different random seeds typically cause the average performance metrics in the evaluation window to move about $0.1 \%$ relative to the reference model's performance metrics. We therefore set our target for normalized $\regret @k$ at $0.1 \%$, since this level of performance loss corresponds to the inherent variance introduced by the random seed choice. This target is marked by black dashed lines in the figures that follow. From now on, unless otherwise specified,  $\regret @k$ refers to the normalized $\regret @k$.

\paragraph{Our most advanced method reaches $\regret @3$ lower than our target level with up to 10 times data reduction, and outperforms both baselines. } Figure~\ref{fig:combined_method} shows the comparison between the combination of our most advanced strategies, which is performance-based stopping with stratified prediction on negative sub-sampled data, and two selected baselines, (1)~basic early stopping and (2)~basic sub-sampling, across five experimental setups. Each plot corresponds to a particular architecture, e.g., FM, CN, MLP, and MoE, with various hyperparameter configurations detailed in Appendix~\ref{sec:app_exp_details}. The difference between ``FM'' and ``FM v2'' plots is the type of hyperparameters we vary. As part of the stratified prediction, we use trajectory prediction with inverse power laws, here and in the rest of the paper. Our advanced method reduces the data usage by up to $10$ times ($C=0.1$) with $\regret @3$ lower than both the target level $0.1$ and the baselines' lowest $\regret @3$. The negative sub-sampling sub-samples the examples with negative labels while keeping all examples with positive labels. The data reduction amount ($C$) is a function of both the performance-based stopping hyperparameters and the sub-sampling ratio. We specify these details in Appendix~\ref{sec:app_exp_details}.

\begin{figure*}[h]
        \centering        \includegraphics[width=0.8\linewidth]{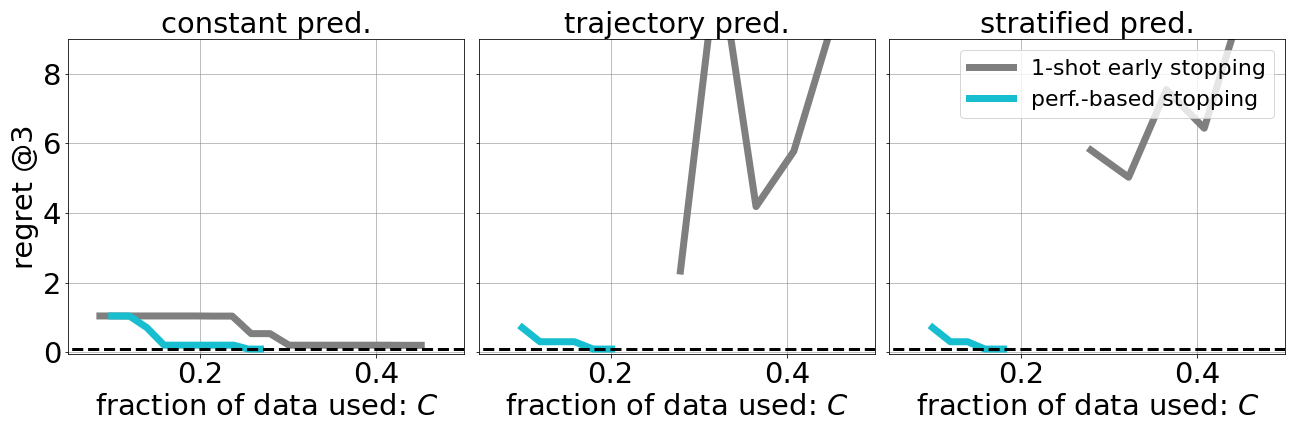}
        \caption{Comparison of one-shot early stopping and performance-based stopping, when used with \textbf{(left)} constant, \textbf{(center)} trajectory, and \textbf{(right)} stratified prediction. }
        \label{fig:early_stopping_vs_SH}
\end{figure*}

\paragraph{Performance-based stopping consistently outperforms one-shot early stopping for any prediction strategy. } Figure~\ref{fig:early_stopping_vs_SH} illustrates that our performance-based stopping needs significantly less data than one-shot early stopping, regardless of the prediction strategy, while still reaching $\regret @3 \leq 0.1$. Here, we show this comparison on MoE experiments with a negative sub-sampling ratio of $0.5$, and provide corresponding results for FM, FM v2, CN, and MLP in Appendix~\ref{sec:app_exp_results}.  

\begin{figure}[h]
  \centering
  \includegraphics[width=0.8\linewidth]{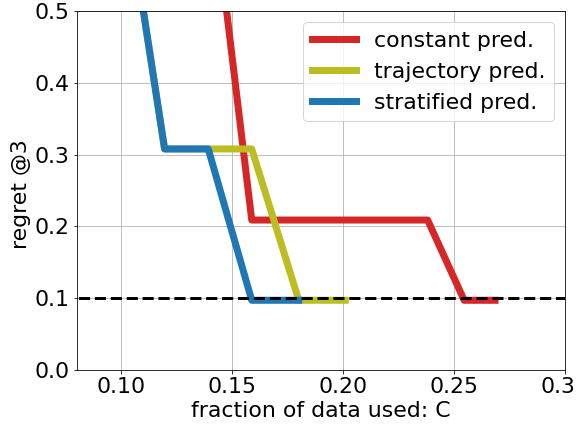}
  \caption{Comparison of prediction strategies. }
  \label{fig:prediction_comp}
\end{figure}

\paragraph{Both trajectory and stratified predictions outperform constant prediction, with stratified prediction providing an additional $10\%$ data reduction over trajectory prediction. } 
Figure~\ref{fig:prediction_comp} shows that both trajectory and stratified predictions reach $\regret @3 \leq 0.1$ level with less than $16\%$ of the training data used, while constant prediction performing significantly worse. We also see that stratified prediction provides $10\%$ additional data reduction gain over trajectory prediction. We show the comparison on MoE here and share similar results for FM, FM v2, CN, and MLP in Appendix~\ref{sec:app_exp_results}.

\paragraph{Additional Results. } We have also explored other choices of laws for trajectory prediction but not observed an improvement over the inverse power law. Moreover, we have tried a ``late starting'' strategy as an alternative to early stopping but have observed similar data reduction vs $\regret @3$ tradeoffs. We present and discuss these additional findings in Appendix~\ref{sec:app_exp_results}.

\subsection{Experiments with an Industrial Web-Scale Advertising System}

To complement the findings on the public benchmark, the core principles of the framework were validated on a real-world, industrial web-scale advertising system. This production system operates on datasets with approximately two orders of magnitude more training data than the Criteo 1TB benchmark, making the cost of a complete methodological ablation study prohibitive.   

Given these constraints, our objective is to validate the applicability and impact of the data reduction framework in this production environment. Therefore, we deploy our performance-based stopping with constant prediction strategy, which the Criteo experiments established as a strong and robust baseline, across several real-world hyperparameter search tasks.   

\begin{figure}[h]
  \centering
  \includegraphics[width=\linewidth]{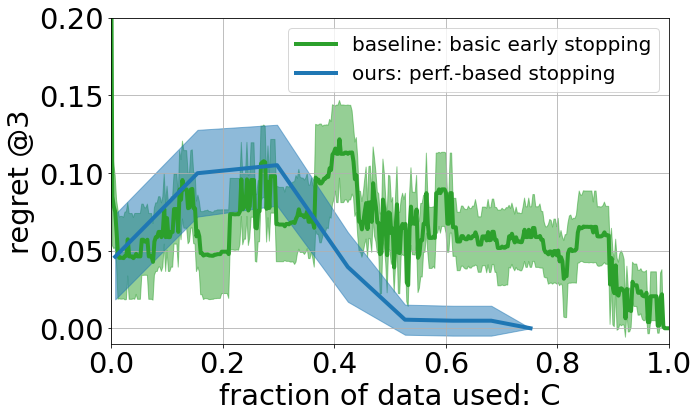}
  \caption{Efficacy of our strategy on industrial web-scale advertising systems. }
  \label{fig:industry}
\end{figure}

Figure~\ref{fig:industry} shows the average cost-regret trade-off curve from these industrial experiments. The results demonstrate that the strategy can reduce computational costs by up to 2$\times$ with very negligible $\regret @3$. A 2$\times$ cost saving is highly significant in practice, as the training cost for a single configuration in such a web-scale system is already substantial. The small standard deviation across multiple search tasks indicates that these efficiency gains are consistent and reliable.   

This result provides powerful, real-world validation that the performance-based stopping with constant prediction method is effective and delivers substantial cost savings at an industrial scale. It confirms that the principles tested on the public Criteo benchmark translate directly to a much larger and more complex production setting.

Furthermore, the comprehensive results on Criteo suggest that even greater gains are achievable. The additional 5$\times$ improvement offered by our more advanced strategies, such as trajectory and stratified prediction, on Criteo points to a significant opportunity for future work. Deploying these more sophisticated strategies within the industrial system could potentially amplify the observed 2$\times$ savings toward the 10$\times$ efficiency gains seen on the public benchmark, creating a compelling case for their integration into production workflows. This narrative connects the scientific depth of the Criteo study with the practical impact demonstrated at industrial scale, using the former to illuminate a path toward even greater efficiencies in the latter.

\section{Discussion \& Conclusion} \label{sec:conclusion}

This work introduces a suite of data reduction and prediction strategies to address the prohibitive cost of hyperparameter search for online learning models in non-stationary environments. The proposed approach is built on a two-stage paradigm that separates the task of identifying promising configurations from the task of training them to their full potential, enabling aggressive and novel cost-saving measures.

The findings are supported by a dual-validation methodology. First, a comprehensive study on the Criteo 1TB dataset (the only suitable public benchmark for this problem) demonstrates the scientific novelty and hierarchical performance of the proposed methods. The performance-based stopping strategy, a generalization of the state-of-the-art SHA algorithm, serves as a strong baseline. Our novel trajectory and stratified prediction strategies, designed explicitly to handle the challenges of distribution shift, are shown to significantly outperform this baseline, achieving up to a 10$\times$ reduction in data requirements. Second, the framework's core principles are validated on a web-scale industrial advertising system, where the generalized baseline alone delivers a 2$\times$ reduction in computational cost -- a result of significant economic value.

While this work achieves significant gains, it also opens avenues for future research. A compelling future direction involves optimizing hyperparameter search directly for live data, which would fundamentally transform how these systems are developed. Furthermore, the successful validation of the baseline method in the industrial system creates a strong motivation for the future deployment of the more advanced trajectory and stratified prediction strategies. The results from the public benchmark suggest that this step could unlock even greater efficiency gains, further bridging the gap between academic research and industrial practice.

\bibliography{main}

\clearpage
\appendix

\section{Additional Experimental Details} \label{sec:app_exp_details}

\subsection{Candidate Configurations} 

In the ``FM'' and ``MoE'' experiments, we vary optimization parameters: learning rate, weight decay, and final learning rate. We sweep through three values for each of these hyperparameters:

\begin{itemize}
    \item learning rate: $[10^{-4}, 10^{-3}, 10^{-2}]$
    \item weight decay: $[10^{-6}, 2 \cdot 10^{-6}, 10^{-5}]$
    \item final learning rate: $[10^{-3}, 10^{-2}, 10^{-1}]$
\end{itemize}

In the ``FM v2'' experiment, in addition to the optimization parameters above, we vary the memory structure of the embeddings. We divide the features into two groups: ``high'' and ``low'' cardinality features and share the embedding tables among them (using hashing). Then, we vary the embedding dimensions as well as the hash buckets for the high and low cardinality features while maintaining a constant training speed and memory footprint. To ensure that all embeddings have the same dimensions in the FM computation, we project them to the same embedding size.

In the ``CN'' and ``MLP'' experiments, in addition to the optimization parameters above, we vary the number of layers and hidden dimensions, respectively:

\begin{itemize}
\item number of layers in CN: $[2,3, 5]$
\item hidden dimensions in MLP: $[(598, 598, 598, 598), (1196, 1196, 1196, 1196)]$
\end{itemize}

\subsection{Compute Resources} 
Without applying our data reduction and prediction strategies, running one configuration over the 24-day Criteo data on 4 TPUv3s takes less than an hour for the architectures and hyperparameters we tried. We manage to reduce this cost up to $10$ times with our strategies. Trajectory prediction requires learning the parameters of a predictive law. We did this using CPUs, with less than an hour for learning the trajectory predictions for each configuration in an experiment of about 100 configurations in total. 

\subsection{Details on Trajectory Prediction}
Trajectory prediction fits a law to the recently observed data and makes predictions for the trajectory. In our experiments, we use the last 3 ``visited'' days in the training data for fitting, by averaging the loss over all the time steps in a day. Then, we use the fitted parameters to predict the loss in the last 3 days, i.e., the evaluation window. In the main body, we use inverse power laws (or \texttt{InversePowerLaw} in Table~\ref{tab:laws_description}). In Section~\ref{sec:other_laws_app}, we explore other choices of laws, listed in Table~\ref{tab:laws_description}, and their combinations. 

\begin{table}[h]
    \centering
    \caption{Other choices of fitting laws for trajectory prediction.}
    \begin{tabular}{cc}
    \toprule
         Law & Formulation (function of $D$) \\
    \midrule
           \texttt{InversePowerLaw} & $E + \frac{A}{D^{\alpha}}$\\
        \texttt{VaporPressure} & $\exp{(A +  \frac{B}{D} + C \cdot \log D)}$\\
         \texttt{LogPower} & $\frac{A}{1 + (\frac{D}{\exp{B}})^\alpha }$\\
        \texttt{ExponentialLaw} & $E - \exp{(-A \cdot D^{\alpha} + B) }$  \\
         \bottomrule
    \end{tabular}
    \label{tab:laws_description}
\end{table}

\subsection{Details on Stratified Prediction}
\begin{figure*}[h]
        \centering
        \includegraphics[width=\linewidth]{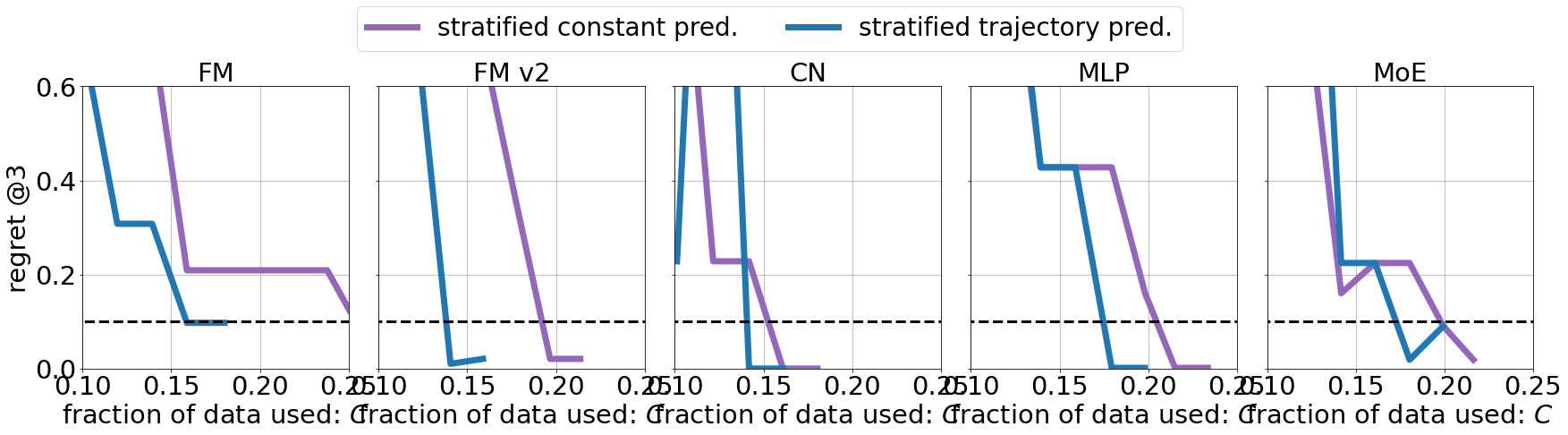}
      
        \caption{Comparison between stratified constant prediction and stratified trajectory prediction. In all other plots in the paper, ``stratified prediction'' refers to stratified trajectory prediction. }
        \label{fig:stratified_choices_app}
\end{figure*}

Stratified prediction can be used with both constant or trajectory prediction. In the main body, we exclusively used it with trajectory prediction as we observe better ranking accuracy. Figure~\ref{fig:stratified_choices_app} shows that stratified trajectory prediction is consistently better than stratified constant prediction across all experiments we consider. Note that in all other plots in the paper, ``stratified prediction'' refers to stratified trajectory prediction. 

\subsection{Details on Performance-Based Stopping}

We provide an outline of our performance-based stopping strategy in Algorithm~\ref{algo:adaptive_stopping}, with two hyperpameters: stopping steps $\mathcal{T}_{\text{stop}}$ and ratio of stopped configurations at any stopping step $\rho$. For all the experiments, we choose $\rho=0.5$. We choose this value since it is a reasonable starting point but it is possible to improve our results with a carefully tuned $\rho$. For $\mathcal{T}_{\text{stop}}$, we fix the number of steps between consecutive stopping times in $\mathcal{T}_{\text{stop}}$ -- i.e., we stop fraction $\rho$ of remaining configurations at equally spaced time steps. By varying the frequency at which we stop the configurations, we obtain different points in the performance-based stopping curves in our plots.

\section{Additional Experimental Results} \label{sec:app_exp_results}

\subsection{One-Shot Early Stopping vs Performance-Based Stopping}
\begin{figure*}[h]
        \centering
        \includegraphics[width=0.8\linewidth]{figures/successive_halving_vs_early_stopping_FM_regret.png}
        \includegraphics[width=0.8\linewidth]{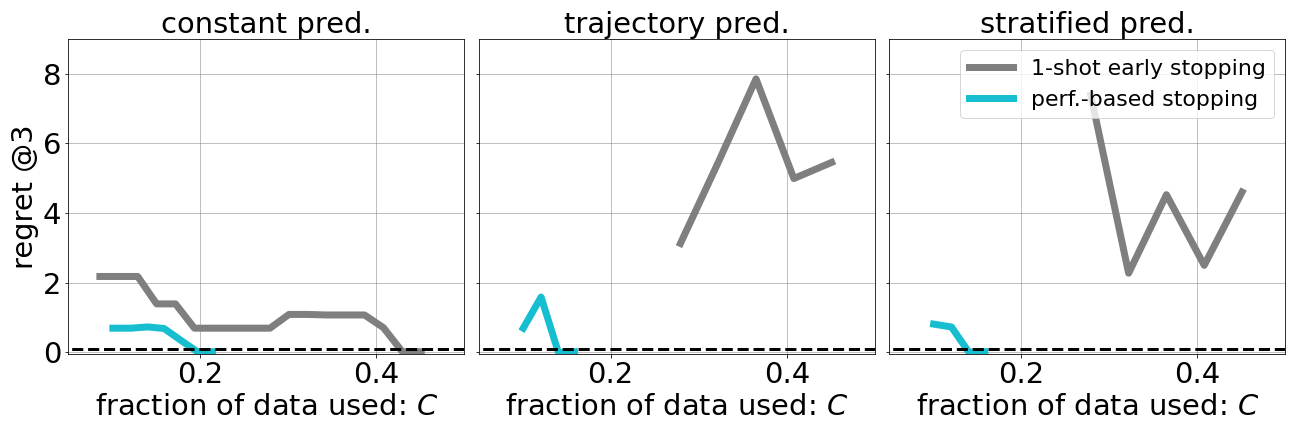}
        \includegraphics[width=0.8\linewidth]{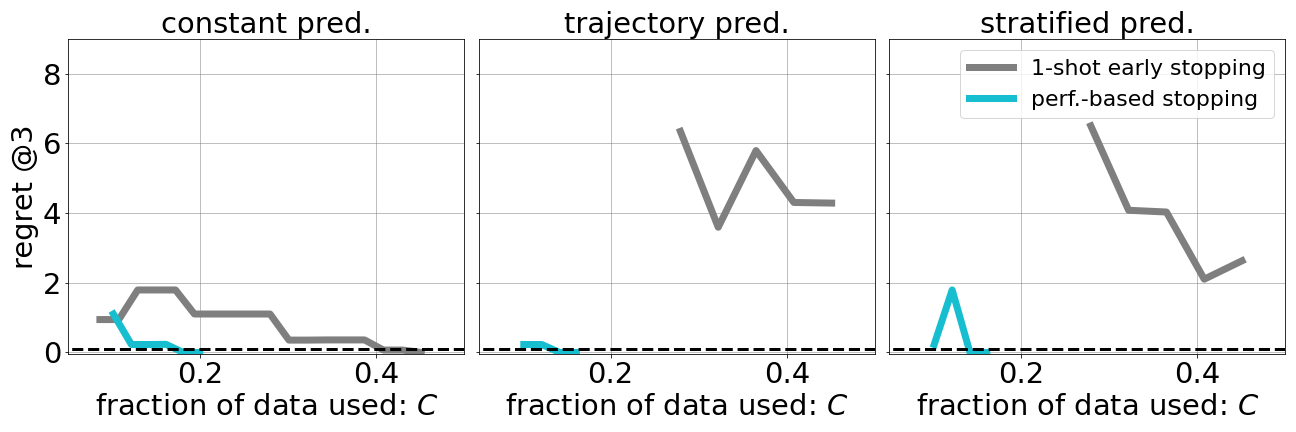}
        \includegraphics[width=0.8\linewidth]{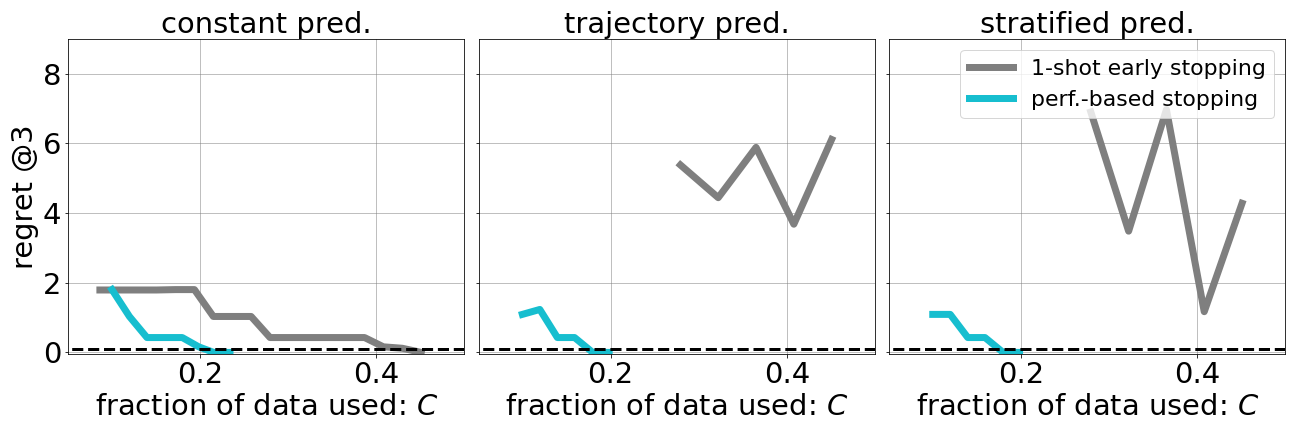}
        \includegraphics[width=0.8\linewidth]{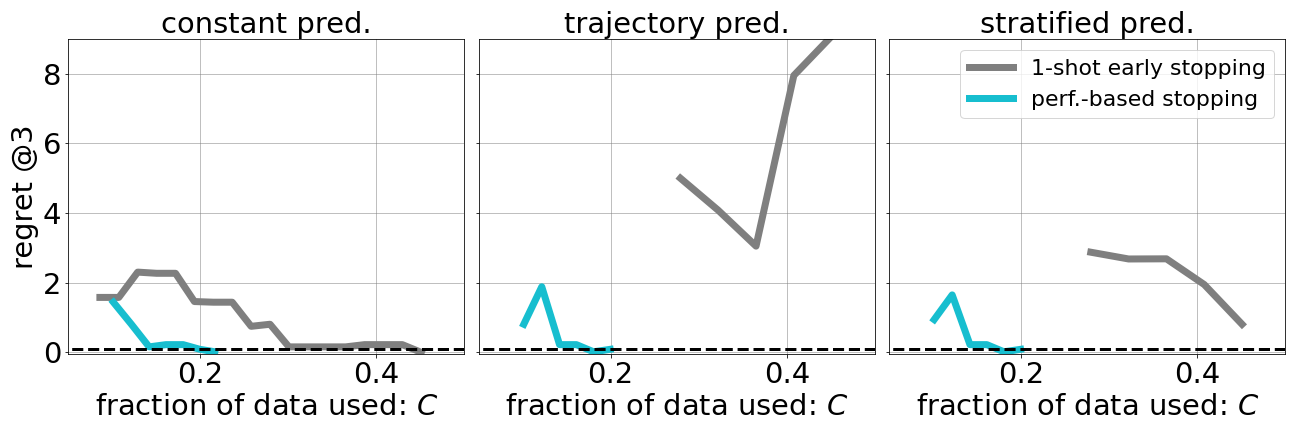}
        \caption{$\regret @3$ comparison of one-shot early stopping and performance-based stopping, when used with \textbf{(left)} constant, \textbf{(center)} trajectory, and \textbf{(right)} stratified prediction. Rows represent, top to bottom: FM, FM v2, CN, MLP, MoE. }
        \label{fig:early_stopping_vs_SH_app_regret}
\end{figure*}

In Figure~\ref{fig:early_stopping_vs_SH}, we demonstrated that performance-based stopping reaches lower $\regret @3$ with significantly smaller amount of data used, compared to one-shot early stopping, and presented results on MoE. We now extend these results to other experiments in Figure~\ref{fig:early_stopping_vs_SH_app_regret}. 

Even though performance-based stopping should not be evaluated with ranking metrics that consider \emph{all} the configurations, like $\fcpr$, as the core idea behind performance-based stopping is to not worry about poor configurations, we still provide a comparison between one-shot early stopping and performance-based stopping based on $\fcpr$ in Figure~\ref{fig:early_stopping_vs_SH_app_per}. We note that this is not a fair comparison since performance-based stopping is actually not expected to rank the poor configurations that stop early on rank accurately. However, Figure~\ref{fig:early_stopping_vs_SH_app_per} shows that the gap between one-shot early stopping and performance-based stopping is so large that the latter reaches much lower $\fcpr$.

\clearpage
\subsection{Comparison between Prediction Strategies}

\begin{figure*}[h]
        \centering
        \includegraphics[width=\linewidth]{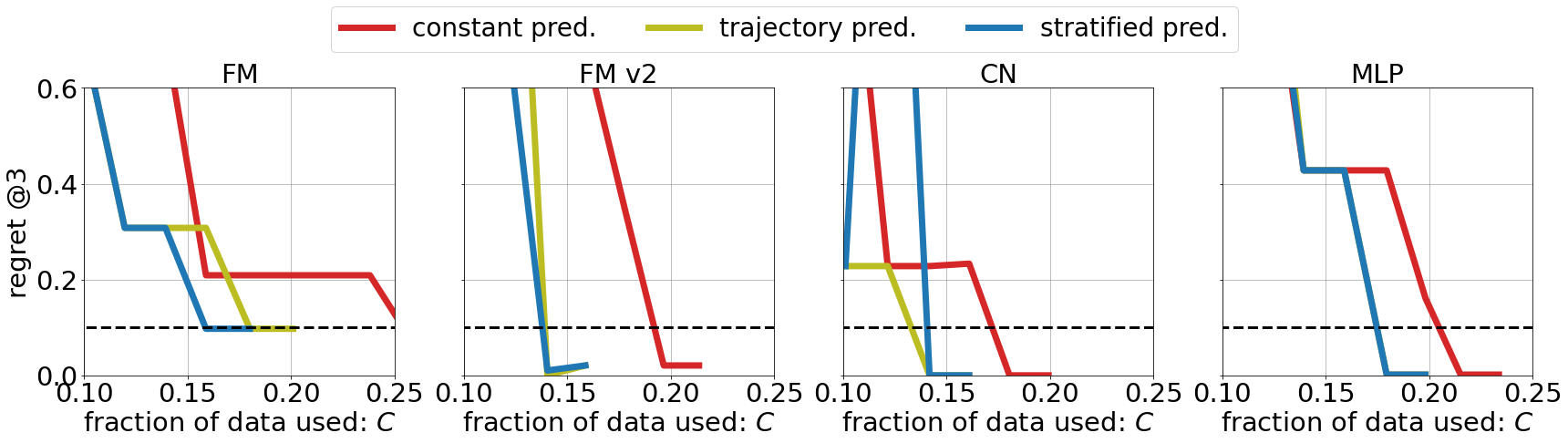}
        \caption{Comparison between our prediction strategies, when used with performance-based stopping. }
        \label{fig:successive_halving_regret_app}
\end{figure*}

We provided a comparison between our prediction strategies on the MoE candidate models in Figure~\ref{fig:prediction_comp}. We extend the same comparison to the other experiments and architectures in Figure~\ref{fig:successive_halving_regret_app}.

\subsection{Other Choices of Scaling Laws} \label{sec:other_laws_app}
\begin{figure*}[h]
        \centering
        \includegraphics[width=0.42\linewidth]{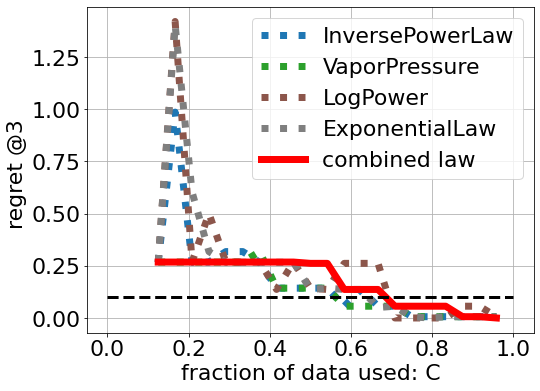}
        \includegraphics[width=0.42\linewidth]{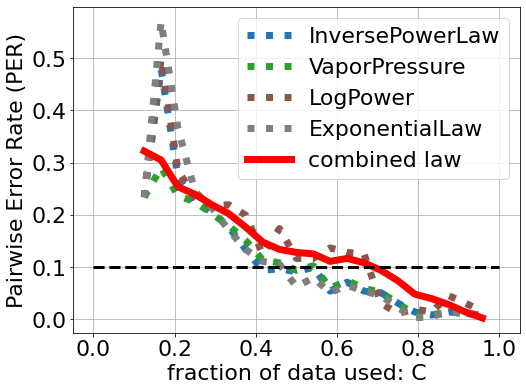}
        \caption{Comparison between different choices of laws for trajectory prediction, with respect to \textbf{(left)  $\regret @3$} and \textbf{(right)} $\fcpr$ . }
        \label{fig:other_lawst_app}
\end{figure*}

Figure~\ref{fig:other_lawst_app} compares different choices of laws, \texttt{InversePowerLaw}, \texttt{VaporPressure}, \texttt{LogPower}, \texttt{ExponentialLaw} defined in Table~\ref{tab:laws_description}, and their weighted combination, for trajectory prediction. It is seen that they all behave similarly and reach the target $\regret @3$ level around the same data use fraction. In the combined law, we learn both the weights and the parameters of each law jointly. 

\subsection{Late Starting vs Early Stopping}
\begin{figure*}[h]
        \centering
        \includegraphics[width=0.8\linewidth]{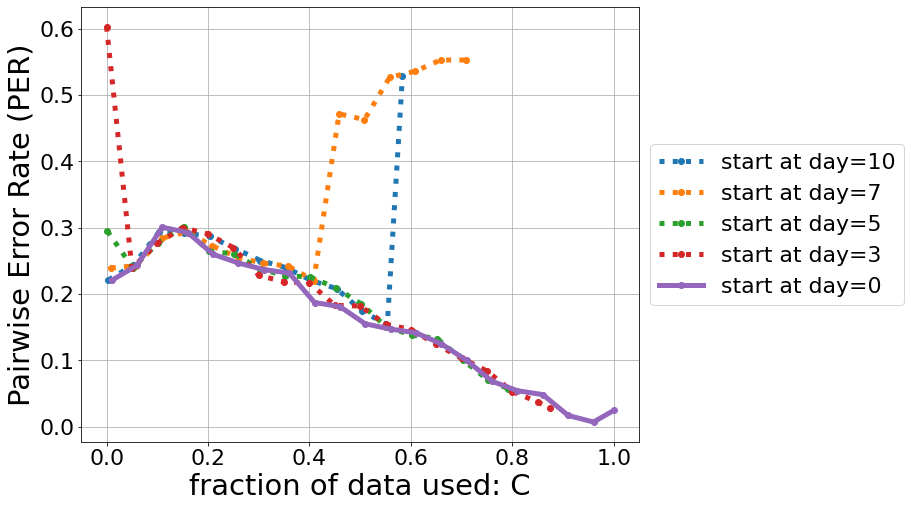}
        \caption{$\fcpr$ comparison between different starting times, when used with one-shot early stopping. This comparison checks whether we could get additional data reduction gain by late starting the configuration runs. However, our ranking predictions with late-started runs provide about the same data reduction vs $\fcpr$ tradeoff.}
        \label{fig:late_starting}
\end{figure*}

Lastly, we explore whether starting come configuration runs later in training data would provide an improvement in the ranking accuracy vs data reduction tradeoff. Figure~\ref{fig:late_starting} shows how one-shot early stopping behaves when the configurations start training at different days. Start at day$=0$ (\textcolor{purple}{purple}) corresponds to the standard one-shot early stopping that we have been analyzing. The other curves apply the same algorithm but on the runs that start training after some number of days in training data passes. We see no significant difference among different start days.

\end{document}